\documentclass[journal]{IEEEtran}
\usepackage{lineno}
\usepackage{fancyhdr,bm,subfigure,graphicx, bibentry,changepage,amsmath,enumerate,amsfonts,amssymb,algorithm,algorithmic}
\usepackage{hyperref}
\usepackage{array}
\usepackage{cite}
\usepackage{enumerate}
\usepackage{epstopdf}

\DeclareMathOperator*{\argmin}{argmin}
\modulolinenumbers[5]

\ifCLASSINFOpdf
\else
\fi
\makeatletter

\newcommand{\thickhline}{%
    \noalign {\ifnum 0=`}\fi \hrule height 1pt
    \futurelet \reserved@a \@xhline
}
\newcolumntype{"}{@{\hskip\tabcolsep\vrule width 1.5pt\hskip\tabcolsep}}
\makeatother

\begin{document}

%
\title{Depth Control of Model-Free AUVs via Reinforcement Learning}
%
%
%


\author{Hui~Wu,
        Shiji~Song, \emph{Member~IEEE},
        Keyou~You, \emph{Member~IEEE},
        and Cheng~Wu
\thanks{This work was supported by National Science Foundation of China (41427806).}
\thanks{The authors are with the Department of Automation and TNList, Tsinghua University, Beijing, 100084, China (e-mail: wuhui115199@163.com; shijis@tsinghua.edu.cn; youky@tsinghua.edu.cn; wuc@tsinghua.edu.cn).}        }

\maketitle

\begin{abstract}
In this paper, we consider depth control problems of an autonomous underwater vehicle (AUV) for tracking the desired depth trajectories. Due to the unknown dynamical model of the AUV, the problems cannot be solved by most of model-based controllers. To this purpose, we formulate the depth control problems of the AUV as continuous-state, continuous-action Markov decision processes (MDPs) under unknown transition probabilities. Based on deterministic policy gradient (DPG) and neural network approximation, we propose a model-free reinforcement learning (RL) algorithm that learns a state-feedback controller from sampled trajectories of the AUV. To improve the performance of the RL algorithm, we further propose a batch-learning scheme through replaying previous prioritized trajectories. We illustrate with simulations that our model-free method is even comparable to the model-based controllers as LQI and NMPC. Moreover, we validate the effectiveness of the proposed RL algorithm on a seafloor data set sampled from the South China Sea.
\end{abstract}


%
\IEEEpeerreviewmaketitle

\section{INTRODUCTION}
\IEEEPARstart{A}{utonomous} underwater vehicle (AUV) is a type of self-controlled submarine whose flexibility, autonomy and size-diversity make it advantageous in many applications, including seabed mapping\cite{kenny2003overview}, chemical pluming tracing\cite{farrell2005chemical}, resource gathering, contaminant source localization \cite{liu2017automated}, operation under dangerous environment, maritime rescue, etc. Therefore, the control of AUVs has drawn great attention of the control community. Among many control problems of AUVs, the depth control is crucial in many applications. For example, when performing a seabed mapping, an AUV is required to keep a constant distance from the seafloor.

There are many difficulties for the depth control problems of AUVs. The nonlinear dynamics of AUVs renders bad performances of many linear controllers such as linear quadratic integral (LQI), fixed proportional-integral-derivative (PID) controllers. Even with nonlinear controllers, sometimes it is hard to obtain an exact dynamical model of the AUV in practice. Moreover, the complicated undersea environment brings various disturbances, e.g., sea currents, waves and model uncertainty, all of which increase the difficulty for the depth control.

Most of conventional control approaches mainly focus on solving the control problems of the AUV based on an exact dynamical model. An extended PID controller including an acceleration feedback is proposed for the dynamical positioning systems of marine surface vessels \cite{lindegaard2003acceleration}. The controller compensates the disturbances of the slow-varying forces by introducing a measured acceleration feedback. A self-adaptive PID controller tuned by Mamdani fuzzy rule is used to control a nonlinear AUV system in tracking the heading and depth with a stabilized speed, and outperforms classical tuned PID controllers \cite{Khodayari2015Modeling}.

Other model-based controllers of AUVs include backstepping\cite{wang2011nn}\cite{zhu2012bio}, sliding-mode\cite{Fang2007On}\cite{wen2017fuzzy}, model predictive control\cite{ai2016source}\cite{li2016trajectory}, etc. An adaptive backstepping controller is designed for the tracking problem of ships and mechanical systems which guarantees uniform global asymptotic stability (UGAS) of the closed-loop tracking error \cite{fossen2001theorem}. Combined with line-of-sight guidance, two sliding mode controllers are respectively designed for the sway-yaw-roll control of a ship\cite{Fang2007On}. Model predictive control (MPC) is a control strategy where control input at each sampling time is estimated based on the predictions over a certain horizon\cite{Budiyono2010Model}. In \cite{sutton2000performance}, a controller is proposed by solving a particular MPC problem and controls a nonlinear constrained submarine to track the sea floor.

However, the performance of model-based controllers will seriously degrade if under an incorrect dynamical model. In practical applications, the accurate dynamical models of AUVs are obviously difficult to obtain due to the complex underwater environment. For such case, a model-free controller is required, which is learned by reinforcement learning (RL) in this work.

RL is a dynamic-programming-based solving framework for the Markov decision process (MDP) without the transition model. It has been applied successfully in the robotic control problems, including the path planning for a single mobile robot\cite{konar2013deterministic} or mutirobot\cite{rakshit2013realization}, robot soccer \cite{Riedmiller2009Reinforcement}, biped robot \cite{zhou2003dynamic}, unmanned aerial vehicle(UAV)\cite{Sch2006An}, etc.

In this paper, we propose a RL framework for the depth control problems of the AUV based on the deterministic policy gradient (DPG) theorem and neural network approximation. We consider three depth control problems including the constant depth control, the curved depth tracking and the seafloor tracking according to different target trajectories and information.

The key of applying RL is how to model the depth control problems as MDPs. MDP describes such a process where an agent at some state takes an action and transits to the next state with an one-step cost. In our problems, the definitions of the `state' and `one-step cost' are significant for the performance of the RL. Usually the motions of an AUV are described by six coordinates and their derivatives. It is straightforward to regard the coordinates as the states of MDPs directly. However, we find this scenario not adaptive since the desired depth trajectories are not included and some of the coordinates are periodic, thus design a better state.

Most of RL algorithms usually approximate an value function and an policy function both of which are used to evaluate and generate a policy respectively. The forms of the approximators are determined according to the transition model of the MDP. However, the stochastic nonlinear dynamics and the constrained control inputs in the depth control problems of the AUV result in serious approximation difficulties for RL.
Therefore, we design neural network approximators considering their powerful representation abilities, where it is crucial to design adaptive structures of the networks for our problems.

After designing the networks, we train them based on the sampled trajectories of the AUV, which compensates the model-free limitation. However, the depth control problems require high real-time because of massive consumed energy when controlling the AUV, which means the sampled data may be not sufficient for the online demand. To improve the data efficiency, we proposed a batch-learning scheme through replaying previous experiences.

The main contributions of our paper are summarized as follows:
\vspace{0.2cm}
\begin{enumerate}[\ 1.\ ]
\item We formulate three depth control problems of AUVs as MDPs with adaptive forms of the states and the cost functions.
\item We design two neural networks with specific structures and rectified linear unit (ReLu) activation functions.
\item We propose a batch-gradient updating scheme called prioritized experience replay and incorporate it into our RL framework.
\end{enumerate}
\vspace{0.2cm}

The reminder of the paper is organized as follows. In Section II, we describe the motions of AUVs and the three depth control problems. In Section III, the depth control problems are modeled as MDPs under appropriate forms of states and one-step cost functions. In Section IV,  a RL algorithm based on the DPG is applied to solve the MDPs. In Section V, we highlight several innovative techniques which are proposed to improve the performance of the RL algorithm. In Section VI, the simulations are performed on a classic REMUS AUV and the performances of two model-based controllers are compared with that of our algorithm to validate its effectiveness. In addition, an experiment on a real seafloor data set is performed to show the practicability of our RL framework.

\section{PROBLEM FORMULATION}
In this section, we describe the coordinate frames of AUVs and the depth control problems.

\subsection{Coordinate Frames of AUVs}
\begin{figure}
    \centering
    \subfigure[]{
    \includegraphics[width=3.5cm]{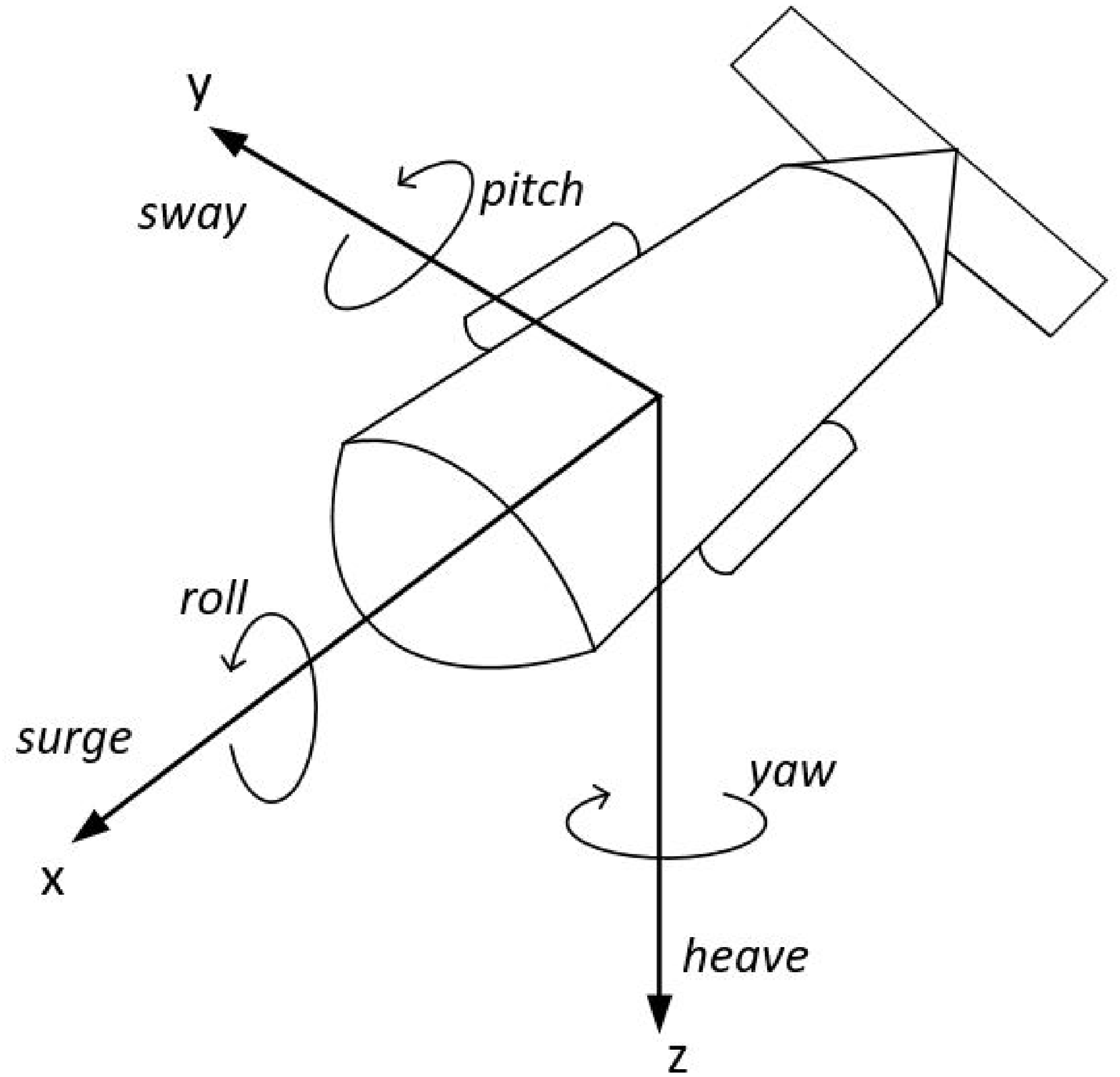}
    \label{fig:6DOFs}}
    \subfigure[]{
    \includegraphics[width=3.5cm]{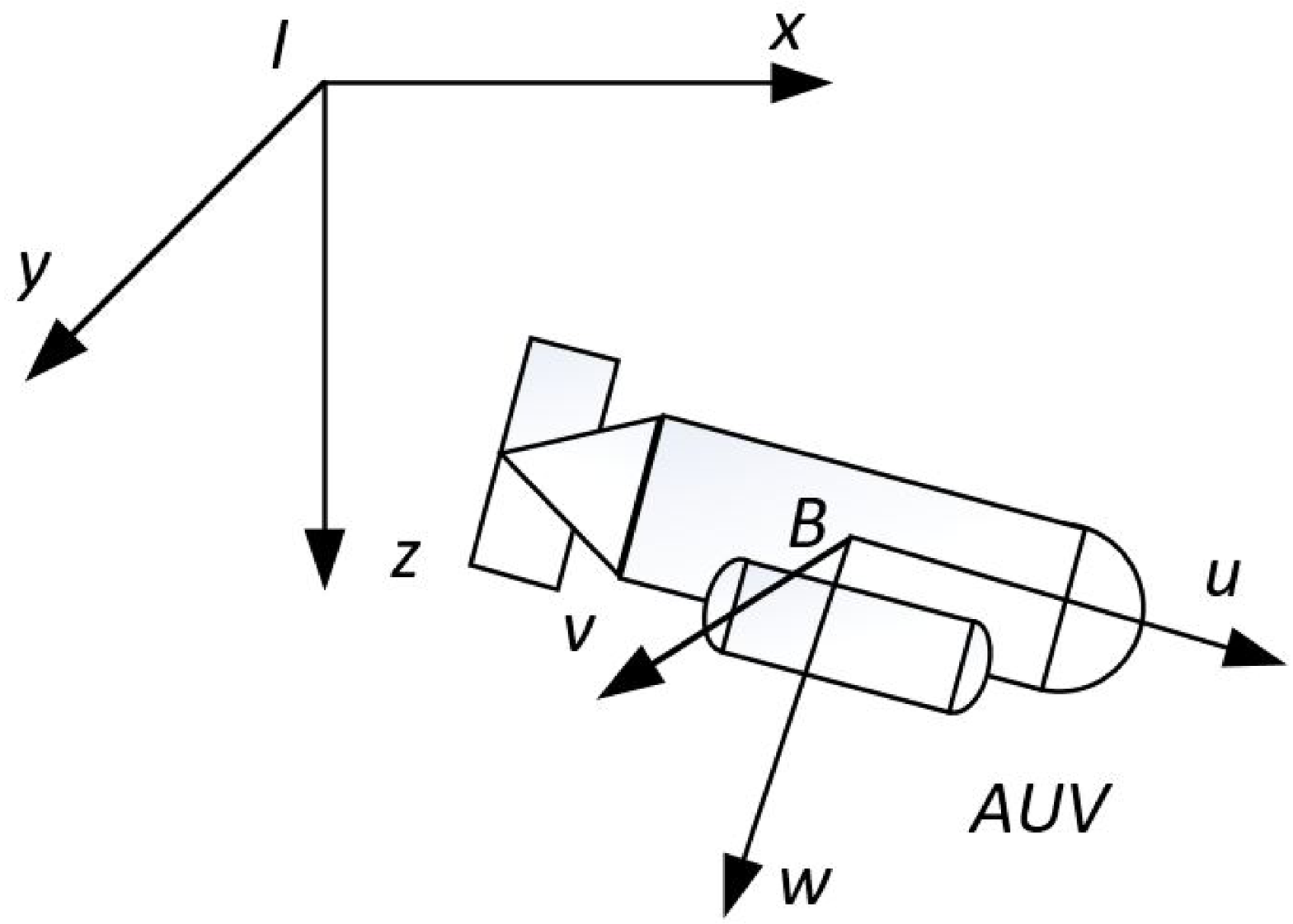}
    \label{fig:coordinate}}
    \label{fig:AUV motion desc}
    \caption{The coordinate description of the AUV. (a)The six DOFs of the motions. (b)The two coordinate frames for describing the AUV motions.}
\end{figure}

The motions of an AUV have six degrees of freedom (DOFs) including \emph{surge}, \emph{sway}, \emph{heave}, which refer to the longitudinal, sideways, vertical displacements, and \emph{yaw}, \emph{roll}, \emph{pitch}, which describe the rotations around the vertical, longitudinal, transverse axises. Fig. \ref{fig:6DOFs} illustrates the details of the six DOFs.

 Correspondingly there are six independent coordinates determining the position and orientation of the AUV. An earth-fixed coordinate frame $\{\bm{I}\}$ is defined for the six coordinates corresponding to the position and the orientation along the x, y and z axes denoted by $\bm{\eta} = [x,y,z,\phi, \theta, \psi]^T$. The earth-fixed frame is assumed to be inertial by ignoring the effect of the earth's rotation. The linear and angular velocities denoted by $\bm{\nu} = [u,v,w,p,q,r]^T$ are described in a body-fixed coordinate frame $\{\bm{B}\}$, which is a moving coordinate frame whose origin is fixed to the AUV. Fig. \ref{fig:coordinate} shows the two coordinate frames and six coordinates.

\subsection{Depth Control Problems}
For simplicity, we just consider the depth control problems of AUVs on the x-z plane in this work, all of which can be easily extended to three-dimensional cases. Thus we only examine the motions on the x-z plane and drop the terms out of the plane. Furthermore, the surge speed $u$ is assumed to be constant. The remaining coordinates are denoted by the vector $\bm{\chi} = [z,\theta, w, q]^T$, including heave position $z$, heave velocity $w$, pitch orientation $\theta$ and pitch angular velocity $q$.

The dynamical equation of the AUV is defined as the follows
\begin{equation}\label{eq:auv model}
\dot{\bm{\chi}} = f(\bm{\chi}, \bm{u}, \bm{\xi})
\end{equation}
where $\bm{u}$ denotes the control vector and $\bm{\xi}$ denotes the possible disturbance.

The purpose of the depth control is to control the AUV to track a desired depth with minimum energy consumption, where the desired depth trajectory $z_r$ is given by
\begin{equation}
z_r = g(x).
\end{equation}

We are concerned with the depth control under three types of situations according to the form and the information about $z_r$:
\vspace{0.2cm}
\begin{enumerate}[\ 1.\ ]
\item \emph{Constant depth control}: The constant depth control is to control the AUV to operate at a constant depth. The desired depth is constant, i.e., $\dot{z}_r = 0$. To prevent the AUV from oscillating around $z_r$, the heading direction is required to keep consistent with the x axis, i.e., the pitch angle $\theta = 0$.
\item \emph{Curved depth tracking}: The curved depth tracking is to control the AUV to track a given curved depth trajectory. The desired depth trajectory is a curve and its derivative $\dot{z}_r$ that describes the tendency of the curve is known. Besides minimizing $|z-z_r|$, the pitch angle is required to keep consistent with the slope angle of the curve.
\item \emph{Seafloor tracking}: Seafloor tracking is to control the AUV to track the seafloor and keep a constant safe distance simultaneously. When tracking the seafloor, sonar-like devices on AUVs can only measure the relative depth $z-z_r$ from itself to the seafloor, thus the slope angle of the seafloor curve is unknown. The missing information makes the control problem more difficult.
\end{enumerate}
\vspace{0.2cm}

\section{MDP MODELING}
In this section, we model the above three depth control problems as MDPs with unknown transition probabilities due to the unknown dynamics of the AUV.

\subsection{Markov Decision Process}
\begin{figure}
\centering
\includegraphics[height = 3.5cm]{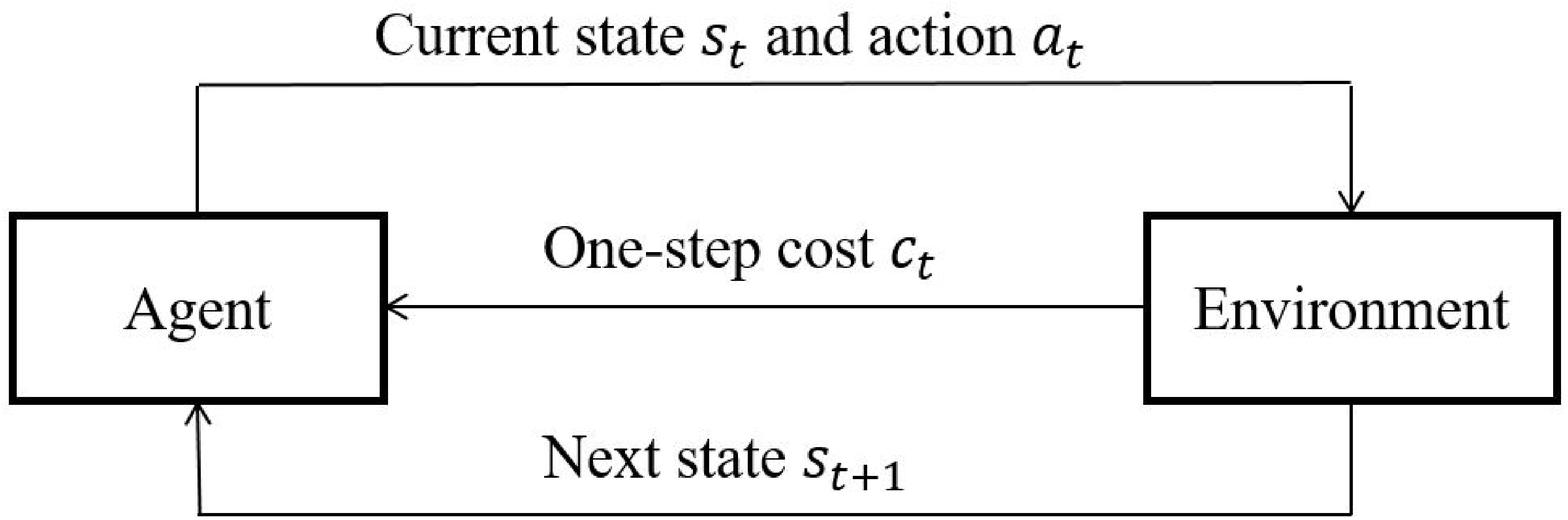}
\caption{The evolution of MDP.}
\label{fig:MDP}
\end{figure}

A MDP is a stochastic process satisfying the Markov property. It consists four components: a state space $\mathcal{S}$, an action space $\mathcal{A}$, an one-step cost function $c(\bm{s},\bm{a}):\mathcal{S}\times \mathcal{A} \rightarrow \mathbb{R}$ and a stationary one-step transition probability $p(\bm{s}_t|\bm{s}_1,\bm{a}_1,\ldots,\bm{s}_{t-1}, \bm{a}_{t-1})$. The Markov property means that current state is only dependent on the last state and action, i.e.,
\begin{equation}
p(\bm{s}_t|\bm{s}_1,\bm{a}_1,\ldots,\bm{s}_{t-1}, \bm{a}_{t-1}) = p(\bm{s}_t|\bm{s}_{t-1}, \bm{a}_{t-1}).
\end{equation}
The MDP describes how an agent interacts with the environment: at some time step $t$, the agent in state $\bm{s}_t$ takes an action $\bm{a}_t$ and transfers to the next state $\bm{s}_{t+1}$ according to the transition probability with an observed one-step cost $c_t = c(\bm{s}_t, \bm{a}_t)$. Fig. \ref{fig:MDP} illustrates the evolution of MDP.

The MDP problem is to find a policy to minimize the long-term cumulative cost function. The policy is a map from the state space $\mathcal{S}$ to the action space $\mathcal{A}$, and can be defined as a function form $\pi: \mathcal{S}\rightarrow \mathcal{A}$ or a distribution $\pi: \mathcal{S}\times \mathcal{A} \rightarrow [0,1]$. Therefore, the optimization problem is formulated as
\begin{equation}\label{eq:J}
\min_{\pi\in \mathcal{P}}{J(\pi)} = \min_{\pi \in \mathcal{P}}{E\left[\sum_{k=1}^{K}{\gamma^{k-1}c_k}|\pi\right]}
\end{equation}
where $\mathcal{P}$ denotes the policy space and $\gamma$ is a discounted factor with $0< \gamma <1$. The superscript $K$ of the summation represents the horizon of the problem.

The definitions of four components of MDPs are essential for the performance of RL algorithms. For the depth control problems of the AUV, the unknown dynamics means the unknown transition probability, and the action corresponds to the control variable, i.e., $\bm{a} = \bm{u}$. Therefore, the key is how to design the states and the one-step cost functions for the control problems.

\subsection{MDP for Constant Depth Control}
The purpose of the constant depth control problem is to control an AUV to operate at a constant depth $z_r$. This task is simple but basic for more complicated cases. We design an one-step cost function of the form
\begin{equation}\label{eq:constant depth cost}
c(\bm{\chi}, \bm{u}) = \rho_1(z-z_r)^2 + \rho_2 \theta^2 + \rho_3 w^2 + \rho_4 q^2 + \bm{u}^T R \bm{u}
\end{equation}
where $\rho_1(z-z_r)^2$ is for minimizing the relative depth, $\rho_2 \theta^2$ for keeping the pitch angle along the x axis and the last three terms for minimizing the consumed energy. The cost function provides a trade-off between different controlling objectives through the coefficients $\rho_1, \rho_2, \rho_3, \rho_4$.

It is intuitive to choose $\bm{\chi}$ which describes the motions of the AUV as the state, but actually this choice is not advisable. Firstly, the pitch angle $\theta$ is an angular variable which cannot be added to the state directly due to its periodicity. A state with $\theta = 0$ and one with $\theta = 2\pi$ are different but actually equivalent with each other. Hence we divide the pitch angle into two trigonometric components $[\cos(\theta), \sin(\theta)]^T$ to eliminates the periodicity.

The second drawback is the absolute depth $z$ in the state. Imagine that if we have controlled an AUV at a specific depth $z_r$, and now the target depth changes to a new depth $z_r'$ that the AUV has never visited. In this case, the control policy for the old $z_r$ does not work for the new depth $z_r'$ because the new states have not been visited before. Thus the relative depth $\Delta z \doteq z-z_r$ is a better choice.

To overcome the above drawbacks, we design the state of the constant depth control as follows
 \begin{equation}\label{constant-state}
 \bm{s} = [\Delta z, \cos(\theta), \sin(\theta), w, q]^T.
\end{equation}

\subsection{MDP for Curved Depth Control}
The curved depth control is to control the AUV to track a given curved depth trajectory $z_r = g(x)$. The state (\ref{constant-state}) is not sufficient for the curved depth control because it does not include the slope angle and its derivative of the curve denoted by $\theta_c$ and $\dot{\theta}_c$. The slope angle $\theta_c$ determines whether the AUV goes uphill or downhill, and $\dot{\theta}_c$ represents the change of rate of $\theta_c$.

\begin{figure}
\centering
\includegraphics[height = 3.5cm]{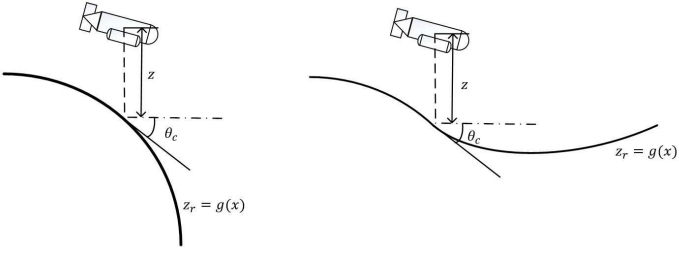}
\caption{Two different situations with different change ratio of $\theta_c$ where the the AUVs in the same state cannot be controlled to track the two curves under the same policy.}
\label{fig:diff theta_c}
\end{figure}

Fig. \ref{fig:diff theta_c} shows two situations where the AUVs in the same state defined according to (\ref{constant-state}) are tracking two curves with different $\dot{\theta}_c$. Obviously they cannot be controlled to track the two curves under the same policy. The failure results from that the AUV cannot anticipate the `tendency' of the curve since the state does not contain $\dot{\theta}_c$.

 In order to add the information about $\dot{\theta}_c$ to the state, we consider the form
\begin{equation}
\dot{\theta_c} = \frac{g''(x)}{[1+g'(x)]^2}\dot{x} = \frac{g''(x)}{[1+g'(x)]^2}(u_0\cos{\theta} + w\sin{\theta})
\end{equation}
where $u_0$ denotes the constant surge speed of the AUV, and $g'(x)$, $g''(x)$ denote the first and second derivative of $g(x)$ with respect to $x$.

Then we consider the derivative of the relative depth $z_{\Delta} \doteq z-z_r$ as follows
\begin{equation}
\begin{split}
\dot{z}_{\Delta} &= \dot{z} - \dot{z}_r\\
&= w\cos{\theta}-u_0 \sin{\theta} + (u_0\cos{\theta} + w\sin{\theta})\tan{\theta_c}\\
&= \frac{w\cos(\theta - \theta_c) - u_0\sin(\theta - \theta_c)}{\cos{\theta_c}} \\
&= \frac{w\cos{\theta_{\Delta} - u_0 \sin{\theta_{\Delta}}}}{\cos{\theta_c}}
\end{split}
\end{equation}
where $\theta_{\Delta} \doteq \theta - \theta_c$ denotes the relative angle between the pitch angle of the AUV and the slope angle of the target depth curve.

Inspired by the form of $\dot{z}_{\Delta}$, we design the following one-step cost function
\begin{equation}\label{eq:curved depth cost}
c(z_{\Delta}, \theta_{\Delta}, w, q, \bm{u}) = \rho_1 z_{\Delta}^2 + \rho_2 \theta_{\Delta}^2 + \rho_3 w^2 + \rho_4 q^2 + \bm{u}^T R\bm{u}
\end{equation}
where $\rho_1 z_{\Delta}^2$ and $\rho_2 \theta_{\Delta}^2$ minimize the relative depth and the relative heading direction respectively. It means that the AUV is controlled to track the depth curve and its tendency simultaneously.

The state for the curved depth control is defined as
\begin{equation}\label{curved-state}
\bm{s} = [z_{\Delta}, \cos{\theta_{\Delta}}, \sin{\theta_{\Delta}}, \cos{\theta_c}, \sin{\theta_c}, \dot{\theta_{\Delta}}, w, q]^T
\end{equation}

\subsection{MDP for Seafloor Tracking}
\begin{figure}
\centering
\includegraphics[width = 8 cm]{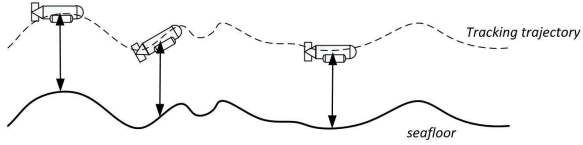}
\caption{The seafloor tracking problem.}
\label{fig:seafloor tracking}
\end{figure}

Illustrated in Fig. \ref{fig:seafloor tracking}, seafloor tracking is to control the AUV to track the seafloor while keeping a constant relative depth. In this case the AUV can only measure the relative vertical distance $\Delta{z}= z - z_r$ from the seafloor by a sonar-like device, but cannot obtain the slope angle and its derivative of the seafloor curve.

Therefore, the state (\ref{curved-state}) is not feasible in this case due to the missing observations of $\theta_c$ and $\dot{\theta}_c$. If we adopt the state defined in (\ref{constant-state}), the problem illustrated in Fig. \ref{fig:diff theta_c} still exists. Actually, this problem is also called ``perceptual aliasing" \cite{Now2012Reinforcement} which means that different parts of the environment appear similar to the sensor system of the AUV. The reason is that the state (\ref{constant-state}) is just a partial observation of the environment. Thus we consider expanding the state to contain more information.

Though not observed by the AUV, the tendency of the seafloor curve can still be estimated by the most recent measured sequence of the relative vertical distances, i.e., $[\Delta{z_{t-N+1}}, \ldots, \Delta{z_{t-1}}, \Delta{z_{t}}]$ where $N$ denotes the window size of the preceding history.

With the same one-step cost function (\ref{eq:constant depth cost}), we define the expanded state for the seafloor tracking problem as
\begin{equation}
\bm{s} = [ \Delta{z_{t-N+1}}, \ldots, \Delta{z_{t-1}}, \Delta{z_{t}}, \cos{\theta}, \sin{\theta}, w,q]^T
\end{equation}

The value of the window size $N$ is significant for the performance of the state, and we determine the best setting for $N$ in the simulation.

\section{SOLVING MDPS OF DEPTH CONTROL VIA RL}
In this section, we adopt the reinforcement learning algorithm to solve the MDPs for the depth control problems of the AUV in Section III.

\subsection{Dynamic Programming}
Here we introduce a classic solving routine for MDPs called \emph{dynamic programming} as the basis of our RL algorithm for the depth control.

We firstly define two types of functions that evaluate the performance of a policy. \emph{Value function} is a long-term cost function defined by
\begin{equation}\label{eq:value function}
V^{\pi}(\bm{s}) = E\left[\sum_{k=1}^{K}{\gamma^{k-1}c_k}|\bm{s}_1 = \bm{s}, \pi\right]
\end{equation}
with a starting state $\bm{s}_1$ under a specific policy $\pi$.
\emph{Action-value} function (also called the Q-value function) is a value function
\begin{equation}
Q^{\pi}(\bm{s},\bm{a}) =  E\left[\sum_{k=1}^{K}{\gamma^{k-1}c_k}|\bm{s}_1 = \bm{s}, \bm{a}_1 = \bm{a}, \pi\right]
\end{equation}
with a chosen starting action $\bm{a}_1$.

Note that the relationship between the long-term cost function (\ref{eq:J}) and value function (\ref{eq:value function}) is given as
\begin{equation}
J(\pi) = \int_{\bm{s}}p_1(\bm{s})V^{\pi}(\bm{s})d\bm{s}
\end{equation}
where $p_1(\bm{s})$ denotes the initial state probability. So the minimization (\ref{eq:J}) is equivalent with the \emph{Bellman optimality equation}
\begin{equation}
\begin{split}
V^*(\bm{s}) = &\min_{\pi\in \mathcal{P}}{V^{\pi}(\bm{s})} \\
= &\min_{\pi\in \mathcal{P}}{\int_{\bm{a}}{\pi(\bm{a}|\bm{s})\cdot }} \\
&{{\left[c(\bm{s}, \bm{a}) + \int_{\bm{s}'}p(\bm{s}'|\bm{s},\bm{a})V^*(\bm{s}')d\bm{s}'\right]d\bm{a}}} \\
\end{split}
\end{equation}

The Bellman optimality equation determines the basic routine of solving the MDP problem comprising two phases known as \emph{policy evaluation} and \emph{policy improvement}. The policy evaluation estimates the value function of a policy $\pi$ as the evaluation of its performance by iteratively using the Bellman equation
\begin{equation}\label{eq:bellman eq}
V^{\pi}_{j+1}(\bm{s}) = \int_{\bm{a}}{\pi(\bm{a}|\bm{s})\left[c(\bm{s},\bm{a}) + \int_{\bm{s}'}p(\bm{s}'|\bm{s},\bm{a})V^{\pi}_{j}(\bm{s}')d\bm{s}'\right]d\bm{a}}
\end{equation}
with an initially supposed value function $V_0^{\pi}(\bm{s})$. The iteration can be performed until the convergence (Policy Iteration) or for fixed steps (Generalized Policy Iteration), or even one step (Value Iteration)\cite{Sutton1998Reinforcement}.

After the policy evaluation, the policy improvement takes place to obtain an improved policy based on the estimated value function by a greedy minimization
\begin{equation}\label{eq:policy improvement}
\pi'(\bm{s}) = \argmin_{\bm{a}\in \mathcal{A}}{\left[c(\bm{s},\bm{a}) + \int_{\bm{s}'}p(\bm{s}'|\bm{s},\bm{a})V^{\pi}(\bm{s}')d\bm{s}'\right]}.
\end{equation}

Both phases are iterated alternatively until the policy converges.

\subsection{Temporal Difference and Value Function Approximation}
The dynamic programming only fits to the MDPs with finite state and finite action spaces under a known transition probability. For the MDPs constructed in Section III, the transition probabilities $p(\bm{s}'|\bm{s},\bm{a})$ are unknown due to the unknown dynamics of the AUV, thus the value function updating (\ref{eq:bellman eq}) cannot be executed. Moreover, the depth control of the AUV is an online control problem which means that the AUV collects sampled trajectories during the tracking process. Thus we adopt a new rule known as \emph{temporal difference} (TD) to update the value function online using newly sampled data.

Suppose that we obtain a transition pair $(\bm{s}_k, \bm{u}_k, \bm{s}_{k+1})$ observed by the AUV at time $k$, then TD gives the updating of the Q-value function as the form \cite{Sutton1998Reinforcement}
\begin{equation}
\begin{split}
Q(\bm{s}_k, \bm{u}_k) \leftarrow &Q(\bm{s}_k, \bm{u}_k) \\
+ &\alpha [c(\bm{s}_k, \bm{u}_k) + \gamma \min_{\bm{u}}{Q(\bm{s}_{k+1}, \bm{u})} - Q(\bm{s}_k, \bm{u}_k)]
\end{split}
\end{equation}
where $\alpha > 0$ is a learning rate.

The TD algorithm updates the map from state-action pairs to their Q values and stores it as a lookup table. However, for the depth control problem, the state consists of the motion vector $\bm{\chi}$ of the AUV and the desired depth $z_r$, while the actions are usually forces and torques of the propellers. All these continuous variables lead to continuous state and action spaces, thus obviously a lookup table is not sufficient.

We represent the map by a parameterized function $Q(\bm{s}, \bm{u}|\bm{\omega})$ and instead update the parameter $\bm{\omega}$ as follows
\begin{equation}\label{eq:update-w}
\begin{split}
&\delta_k = c(\bm{s}_k, \bm{u}_k) + \gamma Q(\bm{s}_{k+1}, \mu(\bm{s}_{k+1}|\bm{\theta})|\bm{\omega}) - Q(\bm{s}_{k}, \bm{u}_{k}|\bm{\omega})\\
&\bm{\omega}_{k+1} = \bm{\omega}_{k} - \alpha \delta_k \nabla_{\bm{\omega}}Q(\bm{s}_k, \bm{u}_k|\bm{\omega})
\end{split}
\end{equation}
where $\mu(\bm{s}_{k+1}|\bm{\theta})$ is a policy function defined in the next subsection.

\subsection{Deterministic Policy Gradient}
The continuous control inputs for the depth control of the AUV result in continuous actions, thus the minimization (\ref{eq:policy improvement}) over the continuous action space is time-consuming if executed every iteration.

Instead, we implement the policy improve phase by the DPG algorithm which updates the policy along the negative gradient of the value function. The DPG algorithm assumes a deterministic parameterized policy function $\mu(\bm{s}|\bm{\theta})$ and updates the parameter $\bm{\theta}$ along the negative gradient of  the long-term cost function of the depth control of the AUV
\begin{equation}
\bm{\theta}_{k+1} \doteq \bm{\theta}_{k} - \alpha \widehat{\nabla_{\bm{\theta}} J(\bm{\theta})}
\end{equation}
where $\widehat{\nabla_{\bm{\theta}} J(\bm{\theta})}$ is a stochastic approximation of the true gradient. The DPG algorithm derives the form of $\widehat{\nabla_{\bm{\theta}} J(\bm{\theta})}$ \cite{Silver2014Deterministic} as follows
\begin{equation}
\widehat{\nabla_{\bm{\theta}}{J(\bm{\theta})}} = \frac{1}{M}\sum_{i=1}^M{\nabla_{\bm{\theta}} \mu (\bm{s}_i|\bm{\theta})\nabla_{\bm{u}_i}Q^{\mu}(\bm{s}_i,\bm{u}_i)}.
\end{equation}
where $Q^{\mu}$ denotes the Q-value function under the policy $\mu(\bm{u}|\bm{s})$ and the approximate gradient is calculated by a transition sequence $[\bm{s}_1, \bm{u}_1,\ldots, \bm{s}_M, \bm{u}_M]$.

In the last subsection, the Q-value function is approximated by a parameterized approximator $Q(\bm{s},\bm{u}|\bm{\omega})$, thus we replace $\nabla_{\bm{u}_i}Q^{\mu}(\bm{s}_i,\bm{u}_i)$ with $\nabla_{\bm{u}_i}Q(\bm{s}_i,\bm{u}_i|\bm{\omega})$.  The approximate gradient is given by
\begin{equation}\label{eq:app-grad}
\widehat{\nabla_{\bm{\theta}}{J(\bm{\theta})}} \thickapprox \frac{1}{M}\sum_{i=1}^M{\nabla_{\bm{\theta}} \mu (\bm{s}_i|\bm{\theta})\nabla_{\bm{u}_i}Q(\bm{s}_i,\bm{u}_i|\bm{\omega})} .
\end{equation}

Note that the approximate gradient in (\ref{eq:app-grad}) is calculated by a transition sequence, which seems to be inappropriate with the online characteristic of the depth control problems. We can still get a online updating rule if setting $M = 1$, but the deviation of the approximation may be amplified. Actually, a batch updating scheme can be performed by sliding the sequence along the trajectory of the AUV.

\vspace{1cm}

We have defined two function approximators $Q(\bm{s},\bm{u}|\bm{\omega})$ and $\mu(\bm{s}|\bm{\theta})$ in TD and DPG algorithms, but do not give the detailed forms of the approximators. Considering the nonlinear and complicated dynamics of the AUV, we constructed two neural network approximators, evaluation network $Q(\bm{s}, \bm{u}|\bm{\omega})$ and policy network $\mu(\bm{s}|\bm{\theta})$ with $\bm{\omega}$ and $\bm{\theta}$ as the weights.

\begin{figure}
\centering
\includegraphics[width = 8.0cm]{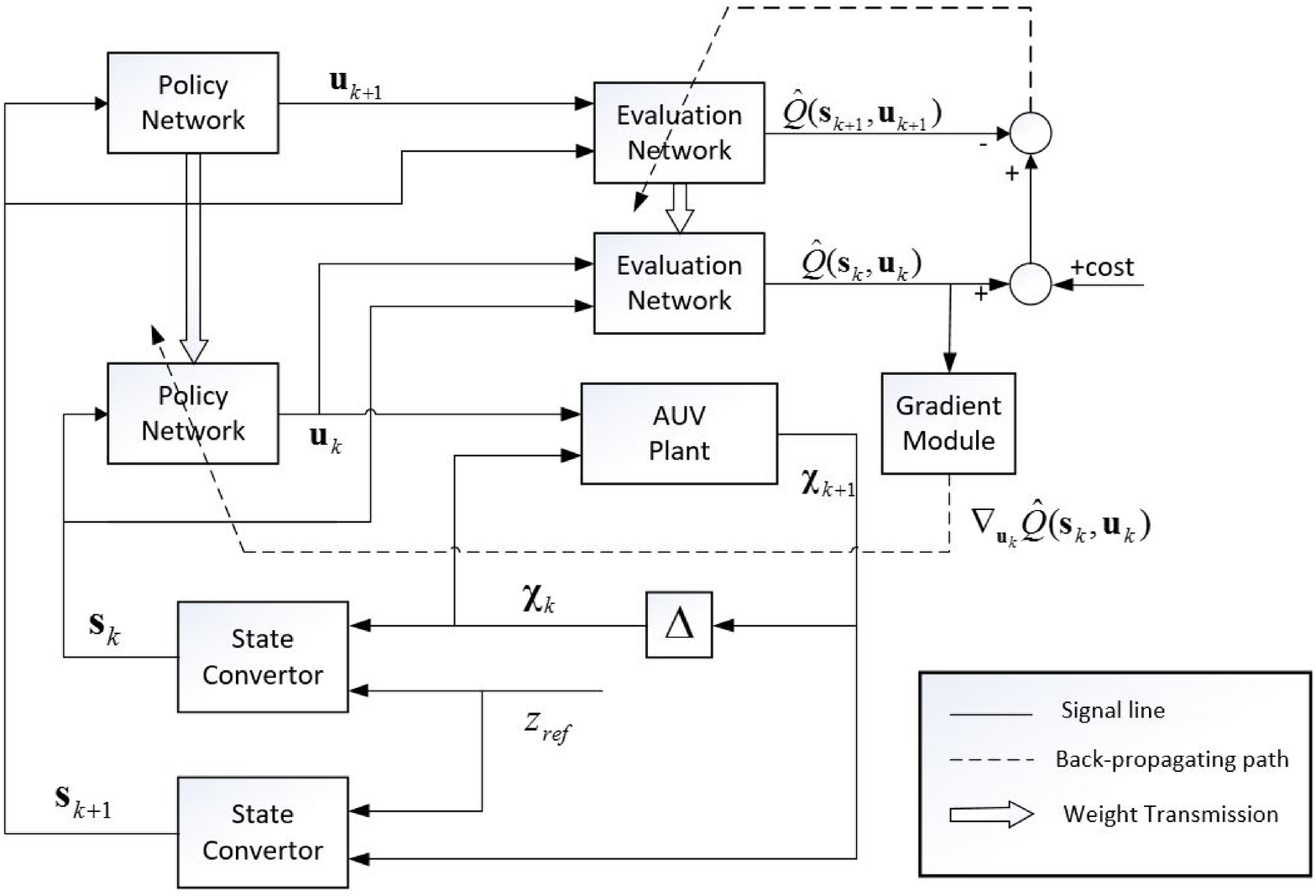}
\caption{The structure diagram of NNDPG.}
\label{fig:struct diag}
\end{figure}

To illustrate the updating of the two networks through TD and DPG algorithms, we present a structure diagram showed in Fig. \ref{fig:struct diag}. The ultimate goal of the RL algorithm is to learn the state feedback controller represented by the policy network. There are two back-propagating paths in our algorithm. The evaluation network is back-propagated by the error between the current Q-value $\hat{Q}(\bm{s}_k,\bm{u}_k)$ and that of successor state-action pair $\hat{Q}(\bm{s}_{k+1},\bm{u}_{k+1})$ plus the one-step cost, which is the idea of the TD algorithm. The output of the evaluation network is then passed to the ``gradient module" to generate the gradient $\nabla_{\bm{u}_k}\hat{Q}(\bm{s}_k, \bm{u}_k)$, which is propagated back to update the policy network through the DPG algorithm. The two back-propagating paths correspond respectively to the policy evaluation and policy improvement phases in dynamic programming. Note that the module ``state convertor" transforms the AUV's coordinates $\bm{\chi}_k$ and reference depth signal $z_{ref}$ into the state $\bm{s}_k$, which denotes the process of state selection illustrated in Section III.

\section{IMPROVED STRATEGIES}
In the last section, we have illustrated a RL framework for the depth control of the AUV, which updates two neural network approximators by iterating the TD and DPG algorithms. Combining the characteristics of the control problems, we further proposed improved strategies from two aspects. Firstly, we design adaptive structures for the neural networks according to the physical constrains when controlling the AUV, where a new type of activation function is adopted. Then we proposed a batch-learning scheme to improve the data-efficiency due to the online features of the depth control problems.

\subsection{Neural Network Approximators}
In Section IV, we define two function approximators $Q(\bm{s},\bm{u}|\bm{\omega})$ and $\mu(\bm{s}|\bm{\theta})$ to approximate Q-value function and policy function respectively, but do not assume the detailed forms of the approximators. Since the dynamics of the AUV is usually nonlinear and complicated, we constructed two neural network approximators, evaluation network $Q(\bm{s}, \bm{u}|\bm{\omega})$ and policy network $\mu(\bm{s}|\bm{\theta})$ with $\bm{\omega}$ and $\bm{\theta}$ as the weights.

The evaluation network has four layers with the state $\bm{s}$ and the control variable $\bm{u}$ as the inputs where $\bm{u}$ is not included until the second layer referring to \cite{Lillicrap2015Continuous}. The output layer is a linear unit to generate a scalar Q-value.

The policy network is designed with three layers and generates the control variable $\bm{u}$ with the inputting state $\bm{s}$. Considering the limit power of the propellers of the AUV, the output $\bm{u}$ must be constrained in a given range. Therefore, we adopt tanh units as the activation functions of the output layer. The output of the tanh function in $[-1, 1]$ is scaled to the given interval.

\begin{figure}
\centering
\includegraphics[height = 4.0cm]{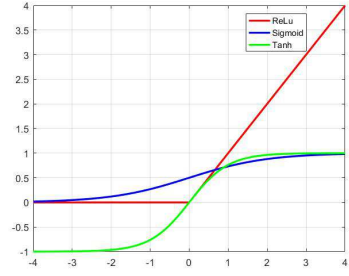}
\caption{The curves of ReLu, Sigmoid and Tanh functions where the ReLu only inhibits changes along the negative x axis and the other two inhibit changes in two directions.}
\label{fig:ReLu}
\end{figure}

Besides the structures of the networks, we also adopt a new type of activation function known as rectified linear unit (ReLu) for the hidden layers. In conventional neural network controllers, the activation functions usually use sigmoid or tanh functions which are sensitive to the changes near zero but insensitive to the changes at large levels. This saturate property brings a `gradient vanish' problem where the gradient of the unit reduces to zero under a large input, which means the large input does not help the training of the network. However, the ReLu function avoids the problem as it only inhibits changes in one direction as illustrated in Fig. \ref{fig:ReLu}. Moreover, the simpler form of the ReLu can accelerate the training of the networks, which just fits the online property of the depth control for the AUV.

\begin{figure}
    \centering
    \subfigure[Evaluation network structure.]{
        \label{fig:evaluation network}
        \includegraphics[height=3.5cm]{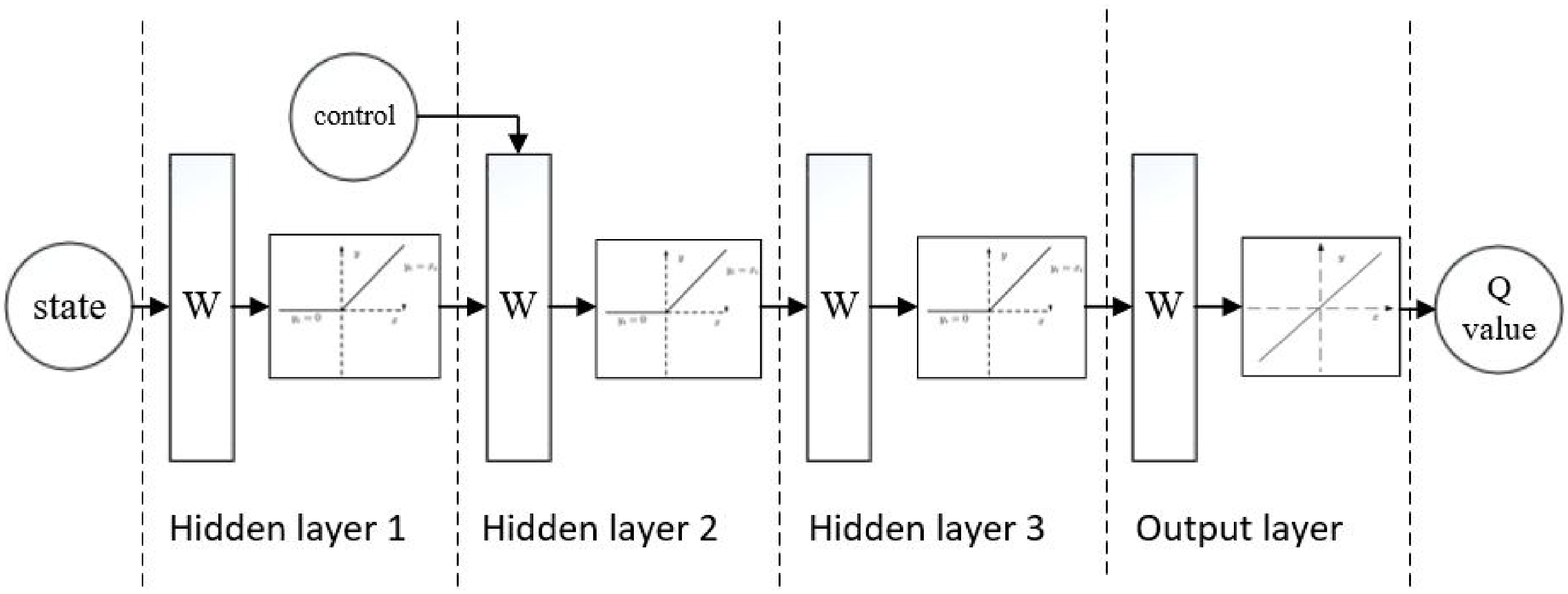}}
    \subfigure[Policy network structure.]{
        \label{fig:policy network}
        \includegraphics[height=3.5cm]{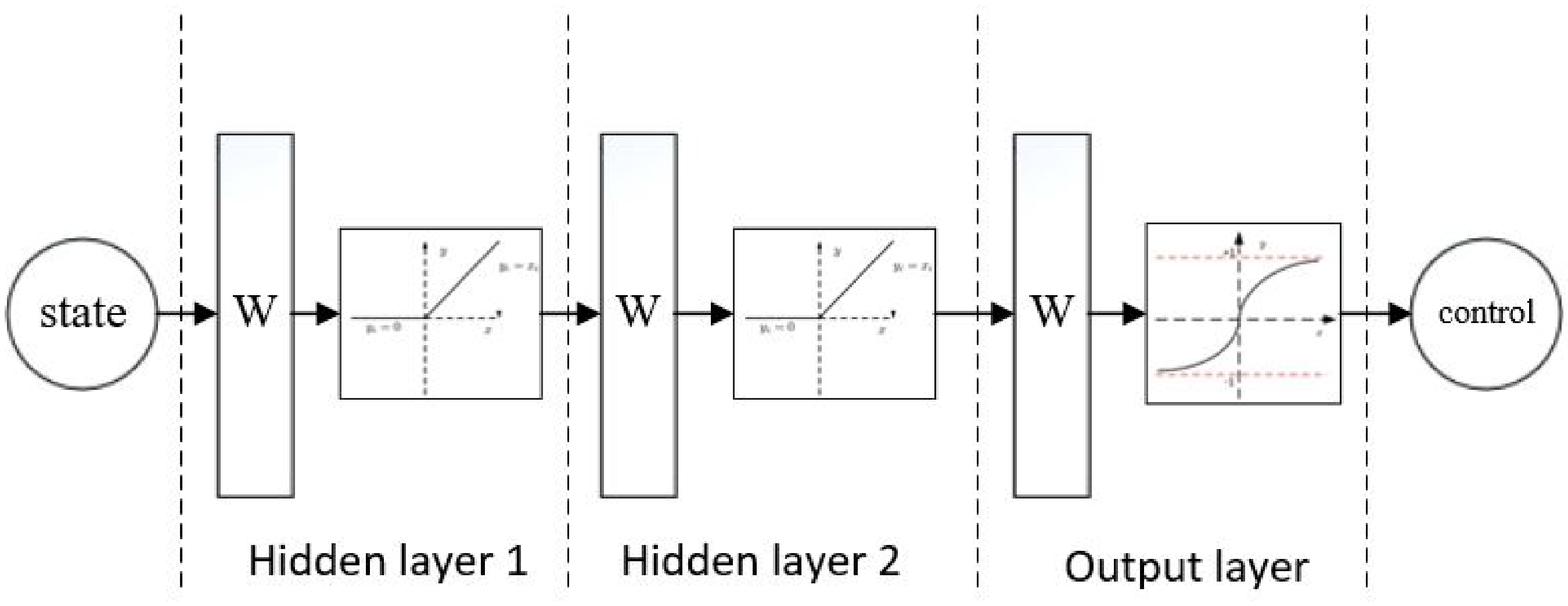}}
    \caption{The structures of two neural networks. The evaluation network has 4 layers with the control input included at the second layer. The policy network has 3 layers with tanh function as the output activation unit.}
    \label{fig:NN structure}
\end{figure}

In the end, we illustrate the full structures of the two networks in Fig. \ref{fig:NN structure}

\subsection{Prioritized Experience Replay}
In this subsection, we consider how to improve the data efficiency for the depth control problems of the AUV. In practice, it consumes large amounts of resources to control the AUV in real undersea environment so that each record along the trajectory is `expensive'. As mentioned, the RL algorithm learns the optimal control policy based on the sampled trajectories data of the AUV. The learning process is necessary but wastes time and resources since the AUV may take some suboptimal even bad actions for interacting fully with the environment. For the depth control of the AUV, we need to shorten the learning process through improving the data efficiency.

We propose a batch learning scheme called \emph{prioritized experience relay} which is an improved version of the \emph{experience replay} proposed by Lin \cite{Lin1992Self}. Imagine the scene of controlling the AUV with the RL algorithm. The AUV observes the subsequent state $\bm{s}'$, one-step cost $c$ in state $\bm{s}$ with control input $\bm{u}$ executed for each time. We call the tuple $(\bm{s}, \bm{u}, \bm{s}', c)$ as an `experience'. Instead of updating the evaluation network and policy network by the newly sampled experience, the prioritized experience replay uses a cache to store all the visited experiences and samples a batch of previous experiences from the cache to update the two networks. The replay mechanism reuses previous experiences as if they were visited newly thus improves the data efficiency greatly.

Actually, not all experiences should be focused equally. If an experience brings minor difference to the weights of the networks, it does not deserve to be replayed since the networks have learned the implicit pattern contained by the experience. Whereas a `wrong' experience should be replayed frequently.

Inspired by the equation (\ref{eq:update-w}), the prioritized experience replay adopts the `error' of the TD algorithm as the priority of an experience, which is given by
\begin{equation}\label{eq:priority}
PRI_k = |c(\bm{s}_k, \bm{u}_k) + \gamma Q(\bm{s}_{k+1}, \mu(\bm{s}_{k+1}|\bm{\theta})|\bm{\omega}) - Q(\bm{s}_{k}, \bm{u}_{k}|\bm{\omega})|.
\end{equation}
The priority measures how probably the experience is sampled from the cache. Therefore, an experience with higher priority makes more differences to the weight of the evaluation network and is replayed with greater probability.

\vspace{1cm}

To sum up, we propose an algorithm called neural-network-based deterministic policy gradient (NNDPG) by combining the above techniques. A more detailed procedure of NNDPG is given in Algorithm \ref{alg:nndpg}.

\begin{algorithm}
\caption{NNDPG for the depth control of the AUV}
\label{alg:nndpg}
\begin{algorithmic}[1]
\REQUIRE ~~\\
Target depth $z_r$, number of episodes $M$, number of steps for each episode $T$, batch size $N$, learning rates for evaluation and policy networks $\alpha_{\bm{\omega}}$ and $\alpha_{\bm{\theta}}$.
\ENSURE ~~\\
Depth control policy $\bm{u} = \mu(\bm{s}|\bm{\theta})$.
\STATE Initialize evaluation network $Q(\bm{s},\bm{u}|\bm{\omega})$ and policy network $\bm{\mu}(\bm{s}|\bm{\theta})$. Initialize the experience replay cache $R$.
\FOR{$i = 1$ to $M$}
\STATE Reset the initial state $\bm{s}_0$.
\FOR{$t = 0$ to $T$}
\STATE Generate a control input $\bm{u}_t = \mu(\bm{s}_t|\bm{\theta}) + \triangle{\bm{u}_t}$ with current policy network and exploration noise $\triangle{\bm{u}_t}$.
\STATE Execute control $\bm{u}_t$ and obtain $\bm{s}_{t+1}$. Calculate one-step cost $c_t$ according to (\ref{eq:constant depth cost}) or (\ref{eq:curved depth cost}).
\STATE Push transition tuple $(\bm{s}_{t}, \bm{u}_t, c_{t+1}, \bm{s}_{t+1})$ into $R$ and calculate its priority according to (\ref{eq:priority})
\STATE Sample a minibatch of $N$ transitions $\{(\bm{s}_{i}, \bm{u}_i, c_{i+1}, \bm{s}_{i+1}) \mid 1 \leq i \leq N \}$ from $R$ according to their priorities.
\STATE Update evaluation network and refresh the priorities of samples. \\
\ \ \ $y_i = c_{i+1} + \gamma Q(\bm{s}_{i+1}, \mu(\bm{s}_{i+1}|\bm{\theta})|\bm{\omega})$\\
\ \ \ $\delta_i = y_i - Q(\bm{s}_{i}, \bm{u}_{i}|\bm{\omega})$\\
\ \ \ $\bm{\omega} \leftarrow \bm{\omega} - \frac{1}{N}\alpha_{\bm{\omega}}\sum_{i=1}^{N}{\delta_i \nabla_{\bm{\omega}}Q(\bm{s}_i, \bm{u}_i|\bm{\omega})}$
\STATE Compute $\nabla_{\bm{u}_i}Q(\bm{s}_i, \bm{u}_i|\bm{\omega})$ for each sample.
\STATE Update policy network by\\
\ \ \ $\bm{\theta} \leftarrow \bm{\theta} - \frac{1}{N}\alpha_{\bm{\theta}}\sum_{i=1}^N{\nabla_{\bm{\theta}}\mu(\bm{s_i}|\bm{\theta})\nabla_{\bm{u}_i}Q(\bm{s}_i, \bm{u}_i|\bm{\omega})}$
\ENDFOR
\ENDFOR
\end{algorithmic}
\end{algorithm}

In the end of this section, it is necessary to discuss the advantages of NNDPG. Firstly, NNDPG does not require any knowledge about the AUV model but can still find control policies whose performance is competitive with the controllers under the exact dynamics of the system. What is more, it greatly improves the data efficiency and the performance by proposing the prioritized experience replay which is firstly used in the RL controller for the control problems of the AUV.

\section{PERFORMANCE STUDY}
\subsection{AUV Dynamics for Simulation}
In this subsection we present a set of explicit dynamical equations of a classical a six-DOF `REMUS' AUV model\cite{Pan2012Depth}, which is utilized to validate our algorithm. However, the experiments can be extended easily to other AUV models since our algorithm is a model-free method.

As mentioned, we only consider the motions of the AUV in x-z plane and assume a constant surge speed $u_0$. The simplified dynamical equations are given by

\begin{subequations}
\begin{equation}\label{eq:dynamic-Z}
\begin{split}
m(\dot{w} - uq - x_G \dot{q} - z_G q^2) = &Z_{\dot{q}}\dot{q} + Z_{\dot{w}}\dot{w} + Z_{uq}uq + Z_{uw}uw \\
&+Z_{ww}w|w| + Z_{qq}q|q|\\
&+(W-B)cos\theta \\
&+ \tau_1 + \triangle \tau_1
\end{split}
\end{equation}
\begin{equation}\label{eq:dynamic-N}
\begin{split}
I_{yy}\dot{q} + m[x_G(uq - \dot{w}) + z_Gwq] = &M_{\dot{q}}\dot{q} + M_{\dot{w}}\dot{w} \\
&+ M_{uq}uq + M_{uw}uw\\
&+ M_{ww}w|w| + M_{qq}q|q|\\
&- (x_G W - x_B B)cos\theta \\
&- (z_G W - z_B B)sin\theta \\
&+ \tau_2 + \triangle \tau_2\\
\end{split}
\end{equation}
\begin{equation} \label{eq:kinematic-Z}
\dot{z} = w cos\theta - u sin\theta
\end{equation}
\begin{equation}\label{eq:kinematic-theta}
\dot{\theta} = q
\end{equation}
\end{subequations}
where $[x_G, y_G, z_G]^T$ and $[x_B, y_B, z_B]^T$ are respectively the centers of gravity and buoyancy; $Z_{\dot{q}}$, $Z_{\dot{w}}$, $M_{\dot{q}}$ and $M_{\dot{w}}$ denote the added masses; $Z_{uq}$, $Z_{uw}$, $M_{uq}$, $M_{uw}$ denote the body lift force and moment coefficients; $Z_{ww}$, $Z_{qq}$, $M_{ww}$, $M_{qq}$ the cross-flow drag coefficients; $W$ and $B$ represent the ROV's weight and buoyancy.
The bounded control inputs $\tau_1$ and $\tau_2$ are propeller thrusts and torques with disturbances $\triangle \tau_1$ and $\triangle \tau_2$ caused by unstable underwater environment. The values of hydrodynamic coefficients are presented in Table \ref{tb:dynamic parameter}

\begin{table}
\centering
\caption{The hydrodynamic parameters of the REMUS AUV}\label{tb:dynamic parameter}
\begin{tabular}{cccc}\hline
Coefficients&Value&Coefficients&Value\\\hline
$m$&30.51&$I_{yy}$&3.45\\
$M_{\dot{q}}$&-4.88&$M_{ww}$&3.18\\
$M_{\dot{w}}$&-1.93&$M_{qq}$&-188\\
$M_{uq}$&-2&$M_{uw}$&24\\
$Z_{\dot{q}}$&-1.93&$Z_{qq}$&-0.632\\
$Z_{\dot{w}}$&-35.5&$Z_{ww}$&-131\\
$Z_{uq}$&-28.6&$Z_{wq}$&-5.22\\\hline
\end{tabular}
\end{table}

\subsection{Linear Quadratic Gaussian Integral Control}
We compare two model-based controllers with the state feedback controller learned by NNDPG. The first is the linear quadratic gaussian integral (LQI) controller \cite{young1972approach} derived from a linearized AUV model.

The nonlinear model of the AUV (\ref{eq:dynamic-Z})-(\ref{eq:kinematic-theta}) can be linearized through SIMULINK linearization mode at a steady state \cite{Khodayari2015Modeling}
\begin{equation}
\begin{split}
w &= w_0 + w'\\
q &= q_0 + q'\\
z &= z_0 + z'\\
\theta &= \theta_0 + \theta'\\
\end{split}
\end{equation}
where $[w', q', z', \theta']^T$ denote the tiny linearization error and the steady state point is set to $[0,0,2.0,0]^T$. The linearized AUV model is derived directly
\begin{equation}
\begin{split}
\dot{\bm{\chi}} &= A \bm{\chi} + B \bm{u} +\bm{\xi}\\
\bm{y} &= C \bm{\chi}
\end{split}
\end{equation}
where coefficient matrices $A, B, C$ are give by
\begin{equation}
\begin{split}
A &= \left(
      \begin{array}{cccc}
        -1.0421 & 0.7856 & 0 & 0.0207 \\
        6.0038 & -0.6624 & 0 & -0.7083 \\
        1.0000 & 0 & 0 & -2.0000 \\
        0 & 1.0000 & 0 & 0 \\
      \end{array}
    \right)\\
B &= \left(
      \begin{array}{cc}
        0.0153 & 0.0035 \\
        -0.0035 & 0.1209 \\
        0 & 0 \\
        0 & 0 \\
      \end{array}
    \right)\
C = \left(
      \begin{array}{cccc}
        0 & 0 & 1 & 0 \\
        0 & 0 & 0 & 1 \\
      \end{array}
    \right)
\end{split}
\end{equation}
and the output $\bm{y} = [z,\theta]^T$.

\begin{figure}
\centering
\includegraphics[height = 3.0cm]{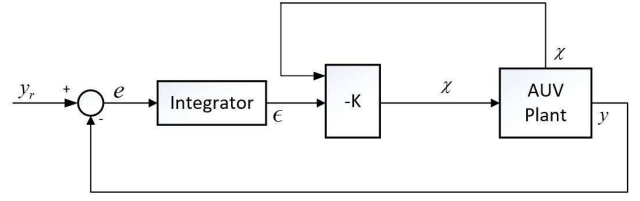}
\caption{The structure diagram of LQI.}
\label{fig:LQI}
\end{figure}

Since the state and output variables are all measurable, a LQI controller is designed to solve the depth control problem for the linearized AUV model showed in Fig. \ref{fig:LQI}. A feedback controller is designed as
\begin{equation}
\bm{u} = K [\bm{\chi}, \bm{\epsilon}]^T = K_{\bm{\chi}} \bm{\chi} + K_{\bm{\epsilon}} \bm{\epsilon}
\end{equation}
where $\bm{\epsilon}$ is the output of the integrator
\begin{equation}
\bm{\epsilon}(t) = \int_0^t{(\bm{y}_r - \bm{y})dt}.
\end{equation}

The gain matrix $K$ is obtained by solving an algebraic Riccati equation, which is derived from the minimization of the following cost function
\begin{equation}
J(\bm{u}) = \int_0^t{\left\{\left[
                              \begin{array}{cc}
                                \bm{\chi}& \bm{\epsilon} \\
                              \end{array}
                            \right]
 Q \left[
                                                   \begin{array}{c}
                                                     \bm{\chi} \\
                                                     \bm{\epsilon} \\
                                                   \end{array}
                                                 \right]
 + \bm{u}^T R \bm{u}\right\}}dt.
\end{equation}

\subsection{Nonlinear Model Predictive Control}
The LQI controller is designed based on an linearized AUV model which is an approximation of the original nonlinear model. Therefore, we adopt a nonlinear controller derived from nonlinear model predictive control (NMPC) under an exact nonlinear AUV dynamics. NMPC designs a N-steps cumulative cost function \cite{sutton2000performance}
\begin{equation}
J_k = \frac{1}{2}\bm{x}_{k+N}^T P_0 \bm{x}_{k+N} + \frac{1}{2}\sum_{i=0}^{N-1}(\bm{x}_{k+i}^T Q \bm{x}_{k+i} + \bm{u}_{k+i}^T R \bm{u}_{k+i}).
\end{equation}
For each time step $k$, NMPC  predicts an optimal N-step control sequence $\{\bm{u}_{k}, \bm{u}_{k+1},\ldots,\bm{u}_{k+N-1}\}$ by minimizing the optimization function
\begin{equation}
\begin{split}
&\min_{\{\bm{u}_{k}, \bm{u}_{k+1},\ldots,\bm{u}_{k+N-1}\}}{J_k}\\
s.t.\ \ &\bm{x}_{i+1} = f(\bm{x}_{i}, \bm{u}_{i})\ \ i=k,\ldots,k+N-1,
\end{split}
\end{equation}
 where the count of predicting steps $N$ is also called prediction horizon. NMPC solves the N-step optimization problem by iterating two processes alternatively. The forward process executes the system equation using a candidate control sequence to find the predictive state sequence $\{\bm{x}_{k+i}\}$. The backward process finds the Lagrange Multipliers to eliminate the partial derivative terms of $J_k$ with respect to the state sequence, and then updates the control sequence along the gradient vector. The two processes are repeated until the desired accuracy.

\subsection{Experiment Settings}
In this subsection, we introduce the experiment settings for the simulations.

The LQI and NMPC controllers are implemented on the Matlab R2017a platform using control system and model predictive control toolboxes. As mentioned, the AUV model is linearized by the SIMULINK from the S-function of the AUV dynamics (\ref{eq:dynamic-Z})-(\ref{eq:kinematic-theta}). The NNDPG is implemented by python2.7 on linux system using the Google's open source module Tensorflow.

It should be noticed that all the controllers and models are implemented as discrete time versions with sample time $dt = 0.1 \ s$, although some of them are described under continuous time in the previous sections. For example, the general AUV dynamics (\ref{eq:auv model}) is discretized using the forward Euler formula
\begin{equation}
\bm{\chi}_{k+1} = \bm{\chi}_{k} + dt \cdot f(\bm{\chi}_{k}, \bm{u}_k).
\end{equation}
The sample horizon is set to $T = 100 \ s$.

The disturbance term $\bm{\xi}$ is generated by an Ornstein-Uhlenbeck process \cite{Mazur1959On}
\begin{equation}
\bm{\xi}_{k+1} = \beta (\bm{\mu} - \bm{\xi}_k) + \sigma \bm{\varepsilon}
\end{equation}
where $\bm{\varepsilon}$ is a noise term complying standard normal distribution, and other parameters are set as $\bm{\mu} = \bm{0}$, $\beta = 0.15$, $\sigma = 0.3$. Note that this is a temporal correlated random process.

\subsection{Simulation Results for Constant Depth Control}
\begin{figure}
\centering
\subfigure[]{
    \label{fig:tracking-z}
    \includegraphics[width = 8.0cm]{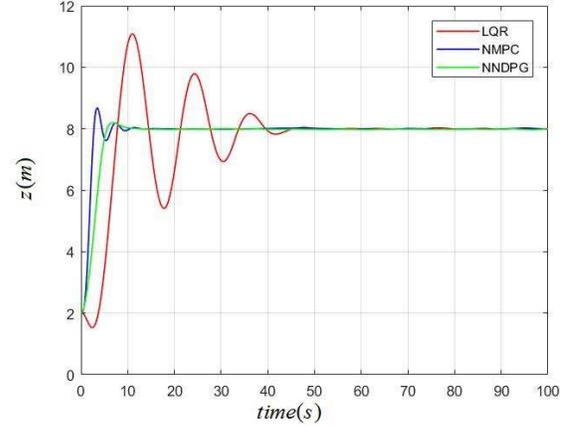}}
\subfigure[]{
    \label{fig:tracking-theta}
    \includegraphics[width = 8.0cm]{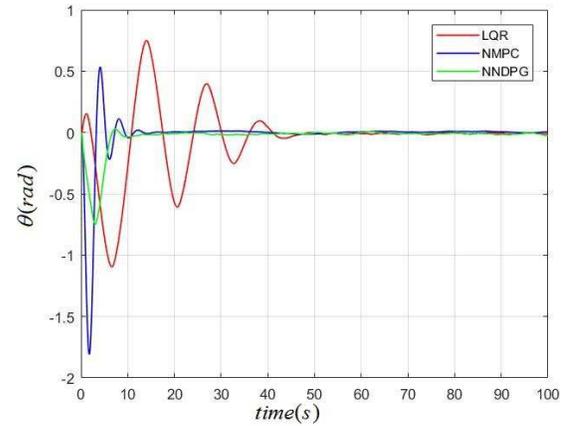}}
\caption{The tracking behaviors of the three controllers for the constant depth control problem where $z_r = 8.0\ m$ and $\theta_r = 0.0 \ rad$. Figures (a) and (b) illustrate respectively the tracking trajectories of the depth $z$ and pitch angle $\theta$.}
\label{fig:track beh}
\end{figure}

Firstly, we compare the performance of NNDPG with those of LQI and NMPC on the constant depth control problem. Fig. \ref{fig:track beh} shows the tracking behaviors of the three controllers from an initial depth $z_0 = 2.0\ m$ to a target depth $z_r = 8.0\ m$. We use three indices, steady-state error (SSE), overshooting and  response time (RT), to evaluate the performance of a controller, whose exact values are present in Table.\ref{tb:perf ind}.

We can find that the LQI behaves worst among the controllers. This result illustrates that the performance of the model-based controller  will deteriorate under an inexact model.

In addition, the simulation shows that the performance of NNDPG is comparable with that of NMPC based on a perfect nonlinear AUV model, and even defeats the latter with a faster convergence speed and a smaller overshooting (bold decimals in Table.\ref{tb:perf ind}). It illustrates the effectiveness of the proposed algorithm under an unknown AUV model.

\begin{table}
\centering
\caption{The performance indexes of three controllers for the constant depth control.}\label{tb:perf ind}
\begin{tabular}{|c"c|c|c|}\hline
Controllers&LQI&NMPC&NNDPG\\\thickhline
$SSE(z)$&$0.0436$&$0.0094$&$0.0191$\\ \hline
$Overshooting(z)$&$3.0849$&$0.6772$&$\bm{0.1945}$\\\hline
$RT(z)$&$42.5$&$7.8$&$\bm{7.3}$\\\thickhline
$SSE(\theta)$&$0.0158$&$0.0065$&$0.0108$\\\hline
$RT(\theta)$&$46.5$&$12.2$&$\bm{11.0}$\\\hline
\end{tabular}
\end{table}

\begin{figure}
\centering
\subfigure[]{
    \label{fig:tracking-tau1}
    \includegraphics[width = 8.0cm]{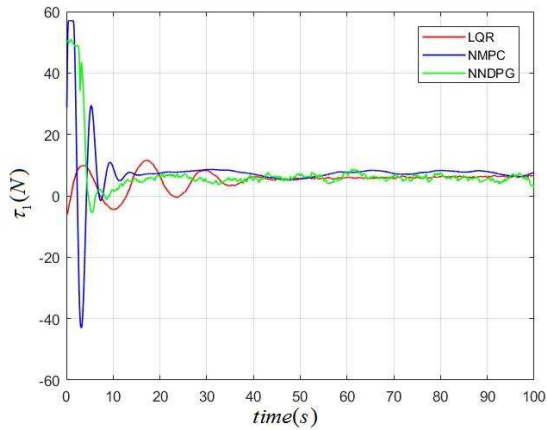}}
\subfigure[]{
    \label{fig:tracking-tau2}
    \includegraphics[width = 8.0cm]{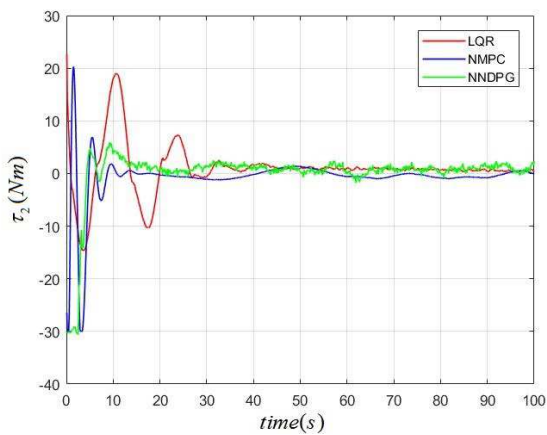}}
\caption{The control sequences of the three controllers for the constant depth control problem. Figures (a) and (b) illustrate respectively the trajectories of $\tau_1$ and $\tau_2$.}
\label{fig:control seq}
\end{figure}

Fig. \ref{fig:control seq} shows the control sequences of the three controllers. The control policy learned by NNDPG changes more sensitively than the other controllers. We explain the phenomenon by the approximation error of the neural network. LQI and NMPG can obtain a smoother control law function since they can access the dynamical equations of the AUV. However, a neural network (policy network) is used to generate the control sequence in NNDPG. Therefore, it can be regarded as a compensation for the unknown dynamical model.

\begin{figure}
\centering
\includegraphics[width = 8.0cm]{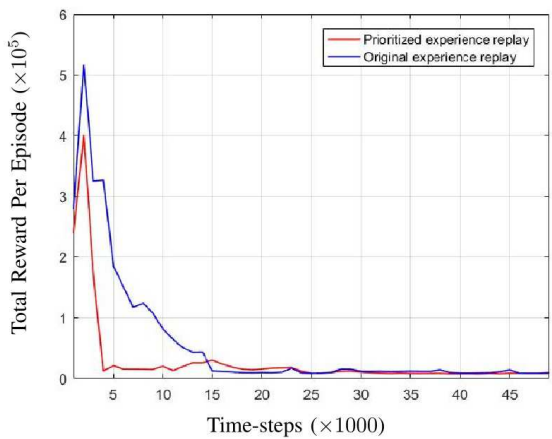}
\caption{Comparison of NNDPG with prioritized experience replay and experience replay.}
\label{fig:pep}
\end{figure}

To validate the improved data-efficiency of the prioritized experience replay, we compare its performance with the original experience replay through the converging process of the total reward, showed in Fig. \ref{fig:pep}. We find that the NNDPG with prioritized experience replay spends less steps in converging than the one with original experience replay, since the former replays previous experiences by an more efficient way.

\subsection{Simulation Results for Curved Depth Control}
\begin{figure}
\centering
\includegraphics[width = 8.0cm]{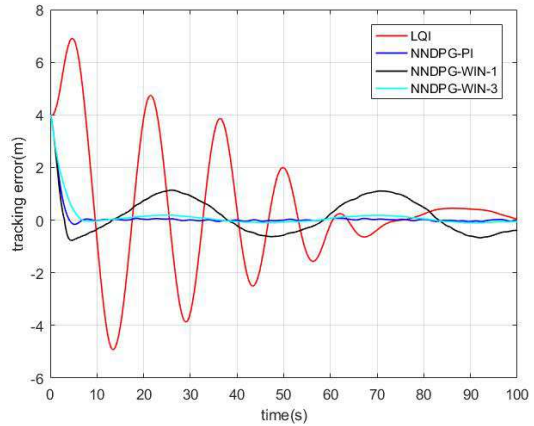}
\caption {The tracking errors of the curved depth control for LQI and different versions of NNDPG.}
\label{fig:tracking-error-sin}
\end{figure}

\begin{figure}
\centering
\includegraphics[width = 8.0cm]{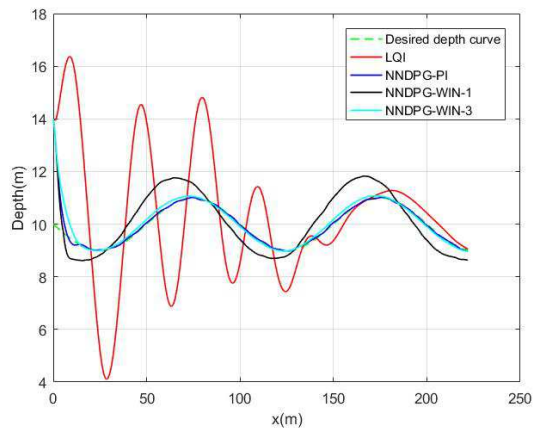}
\caption {The tracking trajectories of the curved depth control for LQI and different versions of NNDPG.}
\label{fig:tracking-traj-sin}
\end{figure}

In this subsection, the AUV is controlled to track a curved depth trajectory. We set the tracked trajectory as a sinusoidal function $z_r = z_{r0} - \sin(\pi/50\cdot x)$ where $z_{r0} = 10\ m$. At first, we assume that the NNDPG has the tendency information about the trajectory including the slope angle and its derivative as the situation of the curved depth control studied in Section III. Then we validate the algorithm under the situation where the slope angle is not measurable. Instead, a preceding historical sequence of the measured relative depths $[\Delta{z_{t-N}}, \Delta{z_{t-N+1}}, \ldots, \Delta{z_{t-1}}, \Delta{z_{t}}]$ is offered where the length of the sequence is called \emph{window size}. The tracking errors and trajectories are showed in Fig. \ref{fig:tracking-error-sin} and Fig. \ref{fig:tracking-traj-sin}, where NNDPG-PI denotes the NNDGP algorithm with the information about the slope angle, and NNDPG-WIN-X denotes the NNDPG involved with $X$ recently measured relative depths.

We can observe that the performance of NNDPG-PI is the best one while NNDPG-WIN-1 including the current relative depth performs much worse,  which validates that the state designed for the constant depth control problem becomes partially observable for the curved depth setting. However, when we add the recent two measured relative depths to the state (NNDPG-WIN-3), the tracking error is much reduced.  The improvement can be explained by the implicit tendency information containing in the latest measurements.

\begin{figure}
\centering
\includegraphics[width = 8.0cm]{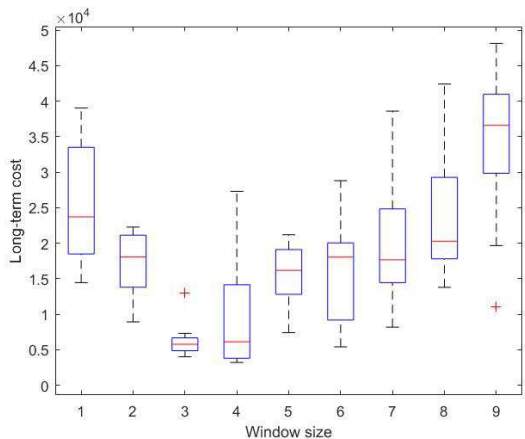}
\caption {The boxplot of long-term cost for different window size.}
\label{fig:long-cost-boxplot}
\end{figure}

To determine the best choice of the window size, we list and compare the performances of NNDPG with different window sizes from 1 to 9. The performances are evaluated according to the long-term cost of one experiment
\begin{equation}
J = \sum_{k=1}^{T}{\gamma^{k-1} c(\bm{s}_k, \bm{u}_k)}.
\end{equation}
Because there are disturbances existing in the AUV dynamics, we make ten experiments for each window size. The results are presented in Fig. \ref{fig:long-cost-boxplot} as a boxplot to show the distributions of the performances.

It is clear to see that the supplement of the past measured relative depths does compensate the missing tendency information of the desired depth trajectory. However, it does not mean more is better. Actually, the records beyond too many steps before are showed to deteriorate the performance because they may disturb the learned policy. From the plot we find the best value of the window size is 3, which corresponds to the minimal mean and variance.

\subsection{Simulation Results of Realistic Seafloor Tracking}
\begin{figure}
\centering
\includegraphics[width = 8.0cm]{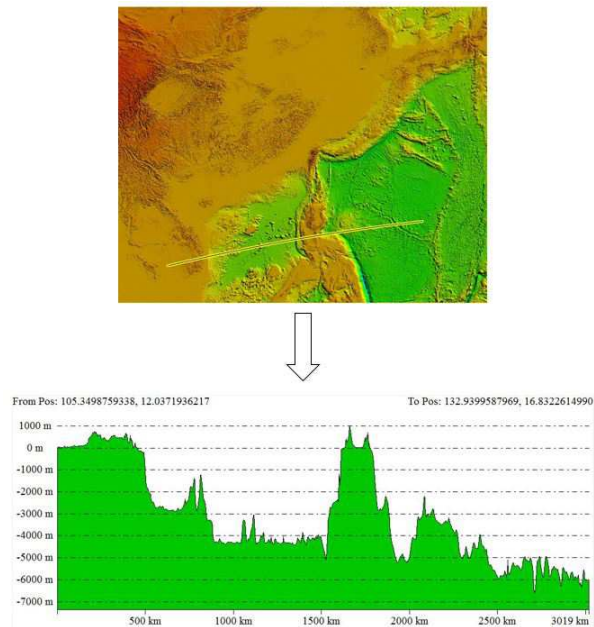}
\caption {The process of sampling the depths data along a preset path.}
\label{fig:seafloor_obt}
\end{figure}

In the end, we test the proposed RL framework for tracking a real seafloor. The data set sampled from the seafloor of the South China Sea at $(23^\circ 06' N, 120^\circ 07' E)$ is provided by the SHENYANG institute of automation Chinese academy of science. We sample the depths along a preset path and obtain a 2D distance-depth seafloor curve showed in the Fig. \ref{fig:seafloor_obt}.

\begin{figure}
\centering
\includegraphics[width = 8.0cm]{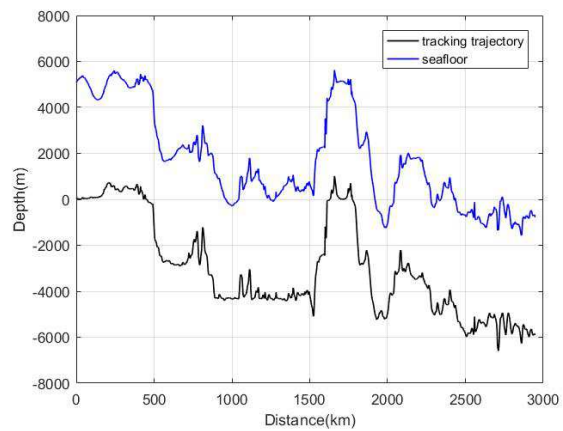}
\caption {The tracking trajectory of NNDPG-WIN-3 and the realistic seafloor.}
\label{fig:real_seafloor_track}
\end{figure}

Our destination is to control the AUV to track the seafloor curve and keep a constant safe distance simultaneously. The tracking trajectory is obtained by the control policy learned by NNDPG-WIN-3 showed in Fig. \ref{fig:real_seafloor_track}. It illustrates that the AUV is controlled to track the rapidly changing seafloor curve with satisfying performance and tracking error.

\section{Conclusion}
This paper has proposed a model-free reinforcement learning framework for the depth control problem of AUVs in discrete time. Three different depth control problems were studied and modeled as a Markov decision process with appropriate forms of the state and one-step cost function. A reinforcement learning algorithm NNDPG was proposed to learn a state-feedback controller represented by a neural network called policy network. The weight of the policy network was updated by an approximate policy gradient calculated based on the deterministic policy gradient theorem, while another evaluation network was used to estimate the state-action value function and updated by the temporal difference algorithm. The alternative updates of two networks composed one iteration step of NNDPG. To improve the convergence, the prioritized experience replay was proposed to replay previous experiences to train the network.

We tested the proposed model-free RL framework on a classical REMUS AUV model and compared the performance with those of two model-based controllers. The results showed that the performance of the policy found by NNDPG can compete with those of the controllers under the exact dynamics of the AUV. In addition, we found that the observability of the chosen state influenced the performance and the recent history could be added to improve the performance.

In the future, we will verify the proposed model-free RL framework on a real underwater vehicle which is a deep-sea controllable visual sampler (DCVS) operating at 6000 meters under the sea level.


%

\appendices


\ifCLASSOPTIONcaptionsoff
  \newpage
\fi

\end{document}